\documentclass[8pt]{article}
\usepackage{graphicx}
\usepackage{pgfplots}
\usepackage{newlfont}
\usepackage[colorlinks,linkcolor=blue,citecolor=blue,
anchorcolor=blue,bookmarksopen,pdfpagetransition={Wipe}]{hyperref}
\usepackage{amsmath,amsthm,amssymb}
\usepackage{subfigure}
\usepackage{graphicx}
\usepackage{amsmath, amsfonts}
\usepackage{dsfont}
\usepackage{booktabs}
\usepackage{fourier}
\usepackage{array}
\usepackage{makecell}
\usepackage{algorithm}
\usepackage{algpseudocode}
\usepackage{lscape}
\usepackage{threeparttable}
\usepackage{hyperref}
\usepackage[T1]{fontenc}
\usepackage[tableposition=top]{caption}
\usepackage{tabularx}

\numberwithin{equation}{section}

\begin{document}
\title{\textbf{Point spread function estimation for blind image deblurring problems
based on framelet transform }}
\author{
Reza Parvaz\footnote{Email: \href{mailto:reza.parvaz@yahoo.com}{reza.parvaz@yahoo.com}, \href{rparvaz@uma.ac.ir}{rparvaz@uma.ac.ir}}}
\date{}
\maketitle
\begin{center}
Department of Mathematics, University of Mohaghegh Ardabili,
56199-11367 Ardabil, Iran.\\
\end{center}
\begin{abstract}
\noindent
One of the most important issues in the image processing is the
approximation of the image that has been lost due to the
blurring process. These types of matters are divided into
non-blind and blind problems. The second type of problem
is more complex in terms of calculations than the first
problems due to the unknown of original image and point spread
function estimation. In the present paper,
an algorithm based on coarse-to-fine iterative by
$l_0-\alpha l_1$ regularization and framelet transform
is introduced to approximate the spread function estimation.
Framelet transfer improves the restored kernel due to the
decomposition of the kernel to different frequencies.
Also in the proposed model fraction gradient operator
is used instead of ordinary gradient operator.
The proposed method is investigated on different kinds of
images such as text,
face, natural. The output of the proposed
method reflects the effectiveness of the proposed algorithm in
restoring the images from blind problems.
\end{abstract}
\vskip.3cm \indent \textit{\textbf{Keywords:}}
Blind deblurring; Framelet; PSF estimation; Natural image; Fractional calculation.
\vskip.3cm

\section{Introduction}
\begin{center}
\begin{figure}[h!]
\centering
\makebox[0pt]{
\includegraphics[width=0.75\textwidth]{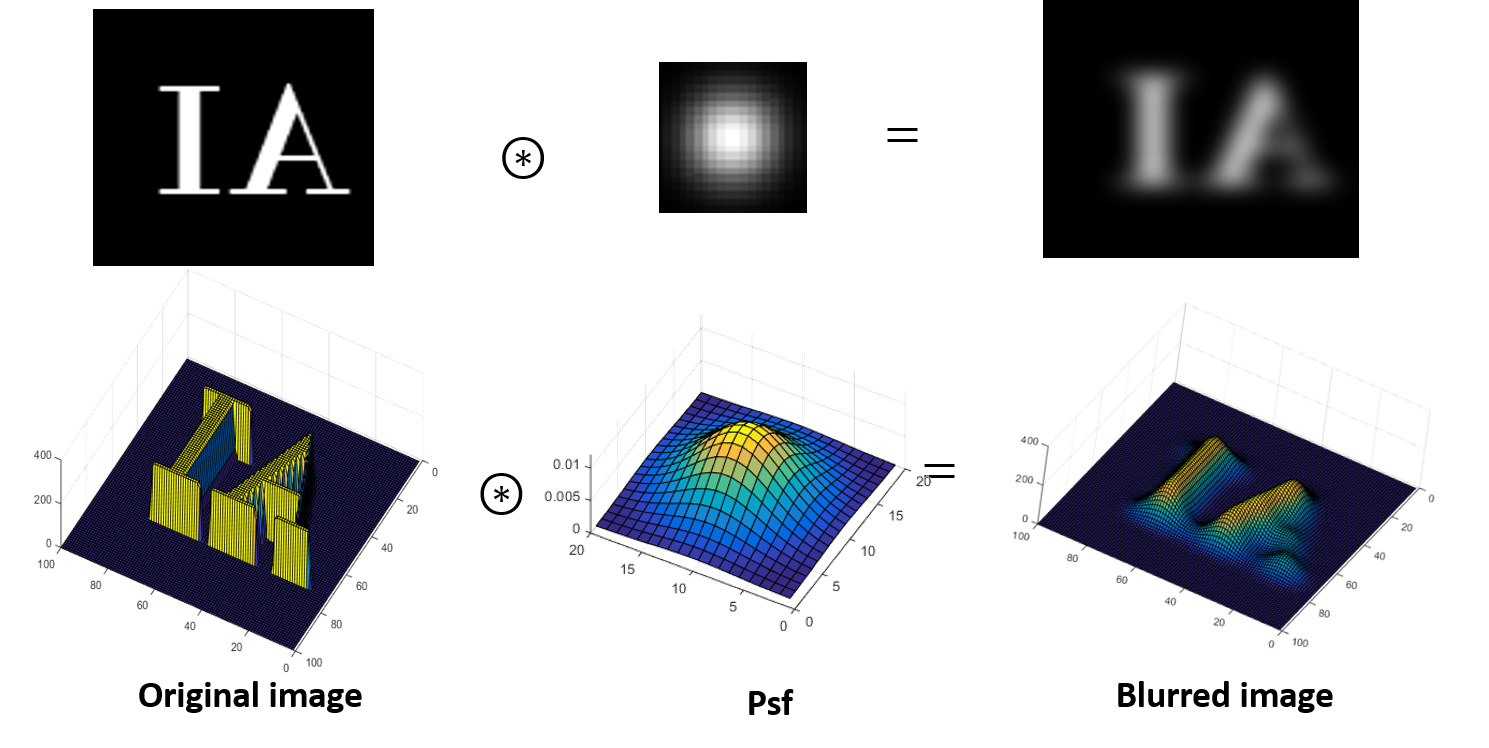}}
\vspace{-0.2cm}
\caption{{\footnotesize
The image blurring without the process of adding noise.
} }
\label{fig01}
\end{figure}
\end{center}

\begin{center}
\begin{figure}[h!]
\centering
\makebox[0pt]{
\includegraphics[width=1.2\textwidth]{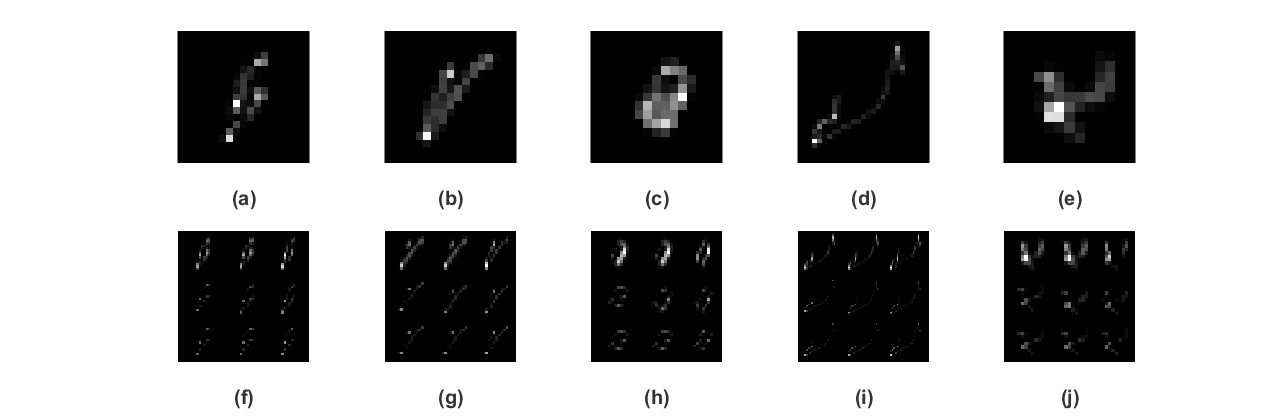}}
\vspace{-0.2cm}
\caption{{\footnotesize
(a-e) Different types of PSF.s from Levin dataset, (f-j) framelet transforms of PSF.s.
} }
\label{fig02}
\end{figure}
\end{center}

One of the most important data in human perception of the environment is
the use of the image taken by the camera. With the growth of social
networks such as Facebook and Instagram, images gained significant
importance. With the spread of smartphones, it has become possible
for everyone to take pictures, and this has led to the spread of
image-based social networks such as Instagram and Pinterest.
Due to the fact that shaking hands or other factors may cause
a decrease in image quality during imaging, so improving the
quality of the image is always one of the most important
challenges for researchers in the field of computer science
and mathematics. Among the most important issues to be
studied in image enhancement is image blurring. These types of
problems can be formulated as 
\begin{align}
y=x\circledast k+\varepsilon,
\end{align}
where $y$ and $x$ in $\mathbb{R}^{n\times m}$ denote blurred and original
images and $\varepsilon$ shows
noise. Also $\circledast$ stands for two dimensional convolution operator
and $k \in \mathbb{R}^{r\times s}$ represents point spread function (PSF),
but why is this name chosen for this matrix ?
The image blurring without the process of adding noise can be seen
in Fig. \ref{fig01}. From this figure, it can be seen that the structure of
the PSF is effective in distributing the blurred image. Considering the
concept of the two dimensional convolution operator,
in fact according to the structure of the PSF and neighboring pixels,
each of the pixels of the original image is changed.
Boundary pixel changes are affected by parts of the outside of the camera
frame that do not appear at the output. This problem has been partially
solved by considering different types of boundary conditions for the
problems. For example, an image taken from the night sky is likely to
have a black area outside the image.
Boundary conditions that are considered according to the structure of
the image are:
zero, periodic, reflexive and anti-reflexive 
boundary conditions \cite{1,4}.
Given that in most cases the problem is removed from the convolution form and
written as a system of linear equations to solve the problems, each of the
boundary conditions is effective on the matrix of coefficients of these
linear equations. For example if boundary condition is considered as zero,
the matrix of coefficients is a block toeplitz with toeplitz
blocks (BTTB) matrix \cite{5}.
The noise factor is usually not considered in the calculations, however,
in some types of blurring problems that have been affected by severe noise,
in addition to the blurring process, the noise-removing process is also performed \cite{6,7}.
According to the number of unknowns, the problem is divided into two types,
the non-blind and blind problems.
In the non-blind problems, blurred image and PSF are known. For example,
an image taken from inside a train traveling at a constant speed has a
motion PSF \cite{8} and is considered the non-blind type of problems.
In the second form, we have only the blurred image, and we have no
information of the PSF.
One of the most effective methods to solve this type of problem is to
use the two-step method for image approximation and PSF approximation,
so that using the iterative method starting from an initial approximation
for the latent sharp image, which is normally blurred image, the PSF is
approximated and then the
approximation of latent sharp image is obtained using the approximated PSF,
and this process continues until we find a suitable results of the PSF and
the latent sharp image.
The coarse-to-fine iterative method is one of the most
widely used methods of this type \cite{9}.
Typically after this step to increase the quality of the
output image,
obtained PSF is used with a non-blind deblurring method.
This step can be selected according to the type of image,
as an example for saturation image, the saturation image
deblurring algorithm \cite{10} can have better performance.
In this type of method, different choices are used for sentences
including the PSF in the proposed model,
for example the $l_0$ norm of the PSF or $l_0$
the gradient of PSF \cite{11}.
Selecting this norm according to the structure of the PSF is
an efficient tool.
With the development of new concepts in mathematics,
a new method is always developed based on these concepts to solve real problems.
The concept of frame is one of the practical concepts
developed from the mathematical analysis and in recent
years has been used in various topics of image processing.
This concept has been extended to various articles on image
deblurring problems, including non-blind and blind problems \cite{12,14}.
In most of these articles, framelet transform is used on the image
and then the algorithm is presented. In the present algorithm, which
is studied in the next section, unlike the most articles, the
penalized term $l_0-\alpha l_1$  is used by framelet transform on the PSF.
The reason for using this method is that images usually have
sparse representations
in the framelet transform domains. And this improves the
PSF approximation because one point of the PSF is approximated at
different frequencies. In order to observe the effect of framelet
transform on the PSF, In Fig. \ref{fig02}, framelet transforms based on
B-spline for
different PSF.s in Levin dataset \cite{15} are shown.
In addition, the method of subtracting two norms is an efficient
method that has been considered in various articles \cite{16,17}. Perhaps
the important problem with this type
of the penalized term is that this method is non-convex.
However, with specific methods, this problem can be turned into a
few convex problems, so this problem can be easily solved.
Another tool used in this paper is to use the fractional derivative. 
This concept is generalized to the ordinary derivative to fractional 
numbers \cite{18}. This tool has been used in recent years in various 
articles on image processing
as \cite{19,20}.
More details of this concept are discussed in the following sections.
\\

\noindent \textit{Notations:} In this paper we use the following notations.
$F(\cdot)$ and $F^{-1}(\cdot)$  are used for fast Fourier transform (FFT) and 
and inverse fast Fourier transform (IFFT),
$\circledast$ and $\langle\cdot , \cdot \rangle$ stand for 
two dimensional convolution operator and inner product.
$H$ shows separable Hilbert space.\\

\noindent\textit{Outline:} The organization of this paper is as follows:
The concepts and tools used in this work are introduced in section \ref{sec2}.
 In Section \ref{sec3}, the
proposed model based on framelet transform and $l_0-\alpha l_1$ regularization
is introduced and numerical algorithms for the latent sharp image and PSF 
approximations are presented.
In Section \ref{sec4}, the proposed algorithm is studied on different
 types of images and different tests are studied to evaluate
the performance on the algorithm.
A summary of the present method is given at the end of the paper in Section \ref{sec5}.

\section{Motivation and preliminaries}\label{sec2}
\begin{figure}[h!]
	\centering
\makebox[0pt]{
	\includegraphics[width=1\textwidth]{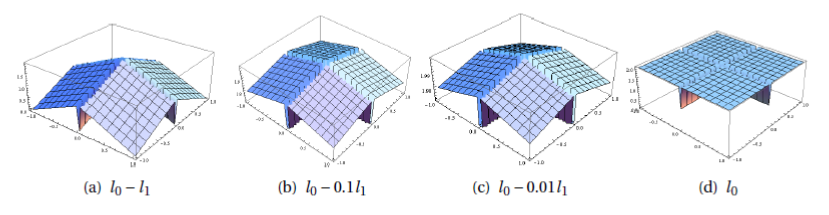}}
	\caption{The plot of $l_0-\alpha l_1$ metric for different values of $\alpha$.}\label{fig03}
\end{figure}

\begin{center}
\begin{figure}[h!]
\centering
\makebox[0pt]{
\includegraphics[width=1\textwidth]{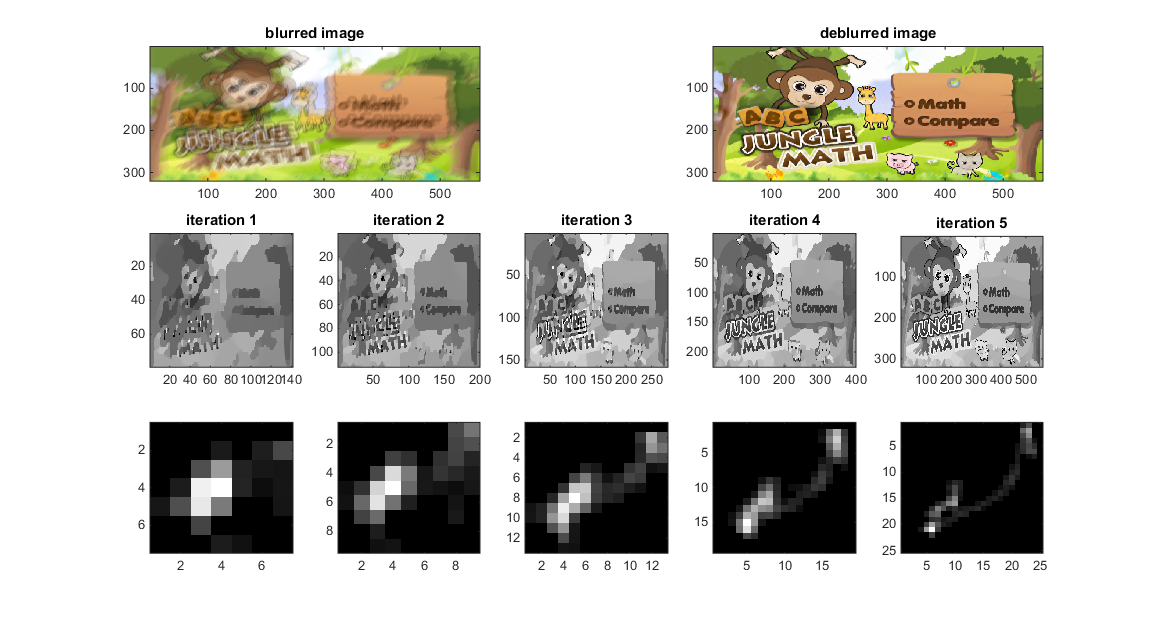}}
\vspace{-0.4cm}
\caption{{\footnotesize
Results of the proposed algorithm in coarse-to-fine steps.
} }
\label{fig04}
\end{figure}
\end{center}

In the following, we explain the details of the proposed model to solve the
blind image deblurring problems.
Different metrics have been used in different papers to approximate the PSF,
and despite having their own advantages, they have some disadvantages.
In \cite{21}, $\|k\|_1$ is used for the PSF, this norm has
a good sparse representation, but it causes the
creation of a PSF with noise. $\| k\|^2_2$
is used in \cite{22}, despite the advantages of this metric such as
convexity and fast speed in calculations and
good noise suppression effect, this method creates
a dense kernel. Another model that is being used to overcome the
disadvantages of previous methods is the use of $\|k\|_0$ for example see
\cite{23}. In \cite{24} this method is improved by adding a second phrase
as $\|\nabla k\|_0$ and this change creates
a PSF with good sparsity with less noise.
Given the structure of the matrix for PSF and considering that most of its 
elements are zero and it can be considered as a sparse matrix. Using the $l_0$ 
norm is a suitable tool to approximate this matrix that
is used in \cite{11}.
One way to approximate the $l_0$ is to use the $l_p,~p\in(0,1)$ because
when $p \rightarrow 0$, $l_p\rightarrow l_0$. These non-convex metrics
are used in \cite{26,28}. The ratio or difference of $l_1$ and $l_2$
is another method to approximate the $l_0$ that had been studied in various articles
\cite{29,31}. Also $l_1-l_2$ had been generalized by $l_1-\alpha l_2$ in \cite{17}.
In the present algorithm, $l_0-\alpha l_1$ metric for $\alpha \in [0,1]$is used 
instead of the $l_0$ metric.
Obviously, when $\alpha \rightarrow 0$, $l_0-\alpha l_1\rightarrow l_0$.
On the other hand, due to the domain of image pixels that are between
zero and one, this metric will always be positive. The graph of this
metric is plotted for $[-1,1] \times [-1,1]$ in Fig. \ref{fig03}.
As can be seen from the figure, this is a non-convex metric.
And that can cause problems in finding a solution.
To solve this problem, the proposed model is divided into
several new sub-models using the methods that are described.
The proposed model, which is introduced, has a suitable behavior in the output
in sparsity and less noise.
The results of the proposed algorithm are presented at each step,
of coarse-to-fine in Fig. \ref{fig04}.
The results show, as the size of PSF increases, the number of pixels with zero value increases.
In the first iteration, approximately $45\%$ of the pixels are equal to zero
and in the last iteration, approximately $14\%$ of the pixels are equal to zero.
Then in the early stages of coarse-to-fine method for kernel approximation, it is
necessary to reduce the effect of $l_0$-norm on the objective function and 
increase the effect by increasing the iteration.
This proposed method is done by reducing the value of a control parameter 
between the two norms i.e., $\alpha$ in $l_0-\alpha l_1$.
On the other hand, the use of framelet transfer makes the edges of the PSF restored properly.
Then the proposed method creates a proper approximations at the early stages of the
PSF in coarse-to-fine algorithm.
The details of the method are presented in the next section.

\subsection{Framelet transform}
One of the fundamental concepts in the signal and image 
processing is the wavelet theory.
In recent years, this discussion has been generalized and 
used in various articles.
The concept of wavelet frames is briefly studied below and 
 reader can find more information about it in \cite{35,36,37}.\\

Let $\langle\cdot , \cdot \rangle $ stands for inner product in separable Hilbert space
$H$
then  a countable set as $X \subseteq H$ is called a tight frame if 
the following condition holds
\begin{align*}
f=\sum_{\psi \in X}\langle \psi, f \rangle \psi ,~~~~\forall f \in  L^2(\mathbb{R}).
\end{align*}
The wavelet system for a given set as $\Psi=\{\psi_1, \ldots, \psi_n\}$ is defined as 
the following collection
\begin{align*}
X(\Psi):=\{ 2^{\frac{j}{2}}\psi_i(2^j \cdot -k)| 1\leq i \leq b; j,k \in \mathbb{Z} \}.
\end{align*}
Also if $X(\Psi)$ be a tight frame in $H$ then
$X(\Psi)$ is named tight wavelet frame and each element of 
$\Psi$ is named a framelet.
In the wavelet theory a compactly supported
scaling function as $\phi$ with refinement mask $h_0$
that $F\big(\phi(2\omega)\big)=F(h_0)F\big(\phi(\omega)\big)$  is used to construct compactly supported wavelet tight frames. Here $F$
stands the Fourier transform. Also a set of framelets for 
a given compactly supported scaling function as $\phi$
to construct a wavelet tight frame are defined by
\begin{align*}
F\big(\phi(2\omega)\big)=F(h_0)F\big(\phi(\omega)\big),~~i=1,\ldots,n,
\end{align*}
where $F(h_i),~i=0,\ldots,n$ are considered as $2\pi$-periodic.
In the proposed algorithm the piecewise linear B-spline framelets \cite{35}
with the refinement mask $F\big(h_0(\omega)\big)=\cos^2(\frac{\omega}{2})$
and two framelet masks $F\big(h_1(\omega)\big)=\frac{-\sqrt{2}i}{2}\sin(\omega)$
and $F\big(h_2(\omega)\big)=\sin^2(\frac{\omega}{2})$
are used. The corresponding lowpass and highpass filters
for these masks are considered as 
$h_0=\frac{1}{4}[1,2,1]$, $h_1=\frac{\sqrt{2}}{4}[1,0,-1]$
and $h_2=\frac{1}{4}[-1,2,-1]$, respectively.
The details of wavelet frame transform can be found in \cite{36,37}. This transform is used in the next section on
the PSF.

\subsection{Fractional calculation}
Fractional calculation is a concept that has been considered
in recent years and has been studied in many different 
fields of science such as physics, mathematics and computer science.
There are different definitions for fractional derivative such as
Riemann-Liouville, Caputo, Caputo-Fabrizio  
and Gr\"{u}nwald-Letnikov (G-L) \cite{18,38}.
As in the use of the ordinary derivative, the discrete form of the derivative 
is considered for image processing.
In the use of the fractional derivative, the discrete form is also efficient. 
And considering that among the definitions of the fractional derivative, 
the discrete form of G-L definition is efficient, so this 
definition is used in image processing \cite{40,42}.\\
 
Let $\Omega \subseteq \mathbb{R}^2$ be a bounded open set then for a real function $u: \Omega \rightarrow \mathbb{R}^2$
the fractional-order gradient is
considered as
\begin{align*}
\nabla^{\lambda} u=\big(\nabla^{\lambda}_h u,\nabla^{\lambda}_v u \big)^T,~~\lambda \in \mathbb{R}^{+}.
\end{align*}
By using G-L definition, we define 
\begin{align*}
&\nabla^{\lambda}_hu(i,j)=\sum^{L-1}_{l=0}(-1)^lC^l_{\lambda}u(i-l,j),\\
&\nabla^{\lambda}_vu(i,j)=\sum^{L-1}_{l=0}(-1)^lC^l_{\lambda}u(i,j-l),
\end{align*}
where
\begin{align*}
C^l_{\lambda}=\frac{\Gamma(\lambda+1)}{\Gamma(l+1) \Gamma(\lambda-l+1)},
\end{align*}
$\Gamma$ stands for gamma function. This definition is used in 
the next section to approximate the PSF.

\section{Proposed model based on $l_0-\alpha l_1$ regularization}\label{sec3}
Details of the proposed model are provided in this section and the 
numerical algorithms for solving the model are discussed.
\subsection{Proposed model}

The proposed objective function of the
deblurring model is introduced as
\begin{align}\label{pm}
\arg \min_{k,x}\|x\circledast k-y\|^2_2+\gamma_1 P^{\sigma}_x+\gamma_2 P^{\alpha}_k,
\end{align}
where
\begin{align*}
&P^{\sigma}_x=\sigma\|x\|_0+\|\nabla x \|_0,\\
&P^{\alpha}_k=\|wk\|_0-\alpha\|k\|_1,
\end{align*}
$\gamma_1,\gamma_2$ and $\sigma$ are the regularization
weights and $\alpha \in [0,1]$; $w$ denotes the framelet transform matrix
such that $w^{T}w=I$ and $\nabla$ stands for gradient operator where
$\nabla x=(\nabla_h x, \nabla_v x)^T$. Also, horizontal and vertical derivatives
are obtained using differential filters $\nabla_h=[1,-1]$
and $\nabla_v=[1,-1]^T$.
As can be seen from the presented model, the part related to
image restoration has not changed with the related papers \cite{11},
but the part of the PSF restoration has changed with terms
described in the previous parts.
This model is non-convex model, therefore, it is necessary to
provide a suitable method for solving. The details of finding
 the answer is introduced in the following subsections.

\subsection{Estimating image with blur kernel}
To approximate the image from the PSF according to the proposed model, the 
following model needs to be solved.
\begin{align}\label{Es11}
\arg \min_{x}\|x\circledast k-y\|^2_2+\gamma_1 \big(\sigma\|x\|_0+\|\nabla x \|_0\big).
\end{align}
The above model is solved in \cite{11}, but in the following, a summary of the method described in that
paper is given. By using auxiliary variables $a$ and $b=(b_h,b_v)$
the problem \eqref{Es11} is rewritten as
\begin{align*}
\arg \min_{x,a,b}\|x\circledast k-y\|^2_2+\beta\|x-a\|^2_2
+\mu_1 \|\nabla x-b \|^2_2+ \gamma_1 \big(\sigma\|a\|_0+\|\nabla b \|_0\big),
\end{align*}
where $\beta$ and $\mu_1$ are positive constant. Based on the new problem
$x$ is obtained by solving
\begin{align*}
\arg \min_{x}\|x\circledast k-y\|^2_2+\beta\|x-a\|^2_2+\mu_1 \|\nabla x-b \|^2_2,
\end{align*}
which the closed-form solution for this subproblem is given by
\begin{align}\label{Es12}
x=F^{-1}\Big( \frac{F^{\ast}(k)F(y)+\beta F(a)+\mu_1 \sum_{i\in \{v,h\}}F^{\ast}(\nabla_i)F(b_i)
}{F^{\ast}(k)F(k)+\beta+\mu_1 \sum_{i \in \{h,v\}}F^{\ast}(\nabla_i)F(\nabla_i)} \Big),
\end{align}
where $\ast$ denotes the complex conjugacy.
Also $a$ and $b$ are obtained by following subproblems
\begin{align*}
&\arg\min_{a} \beta \|x-u\|^2_2+\gamma_1 \sigma \|a\|_0,\\
&\arg \min_{b} \mu_1 \|\nabla x-b \|^2_2+\gamma_1 \|b\|_0.
\end{align*}
These subproblems are pixel-wise minimization problem and the
solutions are given by \cite{43}
\begin{align}\label{Es14}
a = \left\{
\begin{array}{l}
 \begin{array}{*{20}{c}}
   x, & |x|^2 \geq \frac{\gamma_1 \sigma}{\beta},  \\
\end{array} \\
 \begin{array}{*{20}{c}}
   0, & \text{other wise,}  \\
\end{array} \\
 \end{array} \right.
\end{align}
\begin{align}\label{Es15}
b = \left\{ \begin{array}{l}
 \begin{array}{*{20}{c}}
  \nabla x, & |\nabla x|^2 \geq \frac{\gamma_1}{\mu_1},  \\
\end{array} \\
 \begin{array}{*{20}{c}}
   0, & \text{other wise.}  \\
\end{array} \\
 \end{array} \right.
\end{align}
A summary of the described method is given in Algorithm
\textcolor[rgb]{0.00,0.00,1}{1}.

\noindent\line(1,0){280}\\
\vspace{-0.3cm}
Algorithm 1: Image restoration algorithm.\\
\vspace{-0.3cm}
\noindent\line(1,0){280}\\
\begin{algorithmic}
\State \textbf{Input:} Blurred image $y$ and blur kernel $k$.
\State $x \leftarrow y$, $\beta \leftarrow 2\gamma_1 \sigma$.
\Repeat
\State obtain $a$ by \eqref{Es14}.
\State $\mu_1\leftarrow 2\gamma_1$.
\Repeat
\State obtain $b$ by \eqref{Es15}.
\State obtain $x$ by \eqref{Es12}.
\State $\mu_1 \leftarrow 2\mu_1$.
\Until{$\mu_1 > \mu^{max}$}
\State $\beta \leftarrow 2 \beta$.
\Until{$\beta >\beta^{\max}$}
\State \textbf{Output:} Intermediate latent image $x$.
\end{algorithmic}
\line(1,0){280}\\

\subsection{Estimating PSF with image}
In order to approximate the PSF in model \eqref{pm}, it is 
necessary to solve the following problem
\begin{align}\label{Es21}
\arg\min_{k} \|x \circledast k-y\|^2_2+\gamma_2\big(\|wk\|_0-\alpha \|k\|_1\big).
\end{align}
Solving the above problem directly by using the intermediate latent image does not give a good output and
in \cite{9} recommends using the gradient operator to solve the above problem.
Then based on the gradient operator, the following problem is proposed for \eqref{Es21}.
\begin{align*}
\arg\min_{k} \| \nabla x \circledast k-\nabla y\|^2_2+\gamma_2\big(\|wk\|_0-\alpha \|k\|_1\big).
\end{align*}
But in the following, a general model based on the fraction gradient operator
is presented to solve the problem \eqref{Es21} as
\begin{align*}
\arg\min_{k} \| \nabla^{\lambda} x \circledast k
-\nabla^{\lambda} y\|^2_2+\gamma_2\big(\|wk\|_0-\alpha \|k\|_1\big).
\end{align*}
Specifically, when $\lambda$ is equal to $1$, the ordinary model 
by gradient operator is obtained.
By introducing auxiliary variables $c$ and $d$ 
for $wk$ and $k$, respectively, we get
\begin{align*}
&\arg\min_{k,c} \| \nabla^{\lambda} x\circledast k
-\nabla^{\lambda} y\|^2_2+\gamma_2\big(\|c\|_0-\alpha \|d\|_1\big),\\
&\text{s.t.}~c=wk,~d=k
\end{align*}
The augmented form for the above problem can be get as
\begin{align}\label{Es22}
\arg\min_{k,c} \| \nabla^{\lambda} x\circledast k
-\nabla^{\lambda} y\|^2_2+\gamma_2 \|c\|_0+\mu_2\|wk-c\|^2_2-
\gamma_2\alpha\|d\|_1+\mu_3 \|d-k\|^2_2.
\end{align}
To solve the above problem, in the first step, the 
solution of $c$ is studied. By \eqref{Es22}, the subproblem
for $c$ can be written as
\begin{align*}
\arg\min_{c} \gamma_2 \|c\|_0+\mu_2\|wk-c\|_2.
\end{align*}
As discussed in the previous section, this problem is
pixel-wise minimization problem and the solution is obtained as
\begin{align}\label{Es29}
c= \left\{
\begin{array}{l}
 \begin{array}{*{20}{c}}
  wk, & |wk|^2 \geq \frac{\gamma_2}{\mu_2},\\
\end{array} \\
 \begin{array}{*{20}{c}}
   0, & \text{other wise}.\\
\end{array} \\
\end{array}\right.
\end{align}
In the next step, we approximate the value of $k$. For this purpose
by using $\eqref{Es22}$, the following subproblem is obtained
\begin{align}\label{Es24}
\arg\min_{k} \| \nabla^{\lambda} x\circledast k-\nabla^{\lambda} y\|^2_2+\mu_2\|wk-c\|^2_2+\mu_3\|k-d\|^2_2.
\end{align}
By using the optimal condition for \eqref{Es24} and fast Fourier transform,
the closed-form of the final solution for $k$ is written as
\begin{align}\label{Es30}
k=F^{-1}\Big( \frac{\sum_{i\in\{h,v\}}F^{\ast}(\nabla^{\lambda}_i x)
	F(\nabla^{\lambda}_i y)+\mu_3F(d)+\mu_2 F(w^{\ast}c)}
{\sum_{i\in\{h,v\}}F^{\ast}(\nabla^{\lambda}_i x)F(\nabla^{\lambda}_i x)+\mu_3+\mu_2} \Big).
\end{align}
Now, the next subproblem that is needed to find $d$ is studied. This subproblem is written as
\begin{align*}
\arg\min_{d}-\gamma_2\alpha\|d\|_1+\mu_3\|k-d\|^2_2.
\end{align*}
Similar to split Bregman iteration method \cite{44,45}, the following iterative
method is proposed to solve this problem
\begin{align}\label{Es25}
&d^{n+1}=\arg\min_{d}
-\gamma_2\alpha\|d\|_1+\mu_3\|d-k-b^{n}\|^2_2,\\ \label{Es26}
&b^{n+1}=b^{n}+(k-d^{n+1}).
\end{align}
The solution of problem \eqref{Es25}
with condition $|-\frac{\gamma_2\alpha}{\mu_3}|<1 $
 is obtained by using the proximal mapping for $l_1$-norm
as \cite{46}
\begin{align}\label{Es27}
d^{n+1}=k+b^n.
\end{align}
The above processing can be expressed in the following algorithm.
According to the structure of Algorithm \textcolor[rgb]{0,0,1}{2},
the value of $|-\frac{\gamma_2\alpha}{\mu_3}|$ is always lower than one,
therefore, a local minimum value can be found for \eqref{Es25}.

\noindent\line(1,0){280}\\
\vspace{-0.3cm}
Algorithm 2: PSF restoration algorithm.\\
\vspace{-0.3cm}
\noindent\line(1,0){280}\\
\begin{algorithmic}
\State \textbf{Input:} Blurred image $y$ and Intermediate latent image $x$.
\State $b=0, k=0$, $\mu_2 \leftarrow 2\gamma_2$.
\Repeat
\State obtain $c$ by \eqref{Es29}.
\State $\mu_3\leftarrow 2\mu_2$.
\Repeat
\State obtain $d$ by \eqref{Es27}.
\State $b\leftarrow b+k-d$.
\State obtain $k$ by \eqref{Es30}.
\State $\mu_3 \leftarrow 2\mu_3$
\Until{$\mu_3 > \mu_3^{max}$}
\State $\mu_2 \leftarrow 2 \mu_2$.
\Until{$\mu_2 >\mu_2^{\max}$}
\State \textbf{Output:} Intermediate kernel $k$.
\end{algorithmic}
\line(1,0){280}\\

\subsection{Coarse-to-fine framework for PSF}

In the previous subsections, the methods of image restoring by the PSF and
PSF restoring by the image are studied.
Using the above methods directly to restore image information does
not always work. One of the most effective methods that is considered
in various algorithms is the use of an
image pyramid framework in a coarse-to-fine method \cite{9}.
 Due to the fact that sharp edges is effective in approximating the PSF,
 so using the fraction gradient operator improves the kernel.
 This algorithm is given in Algorithm \textcolor[rgb]{0.00,0.00,1.00}{3}.
 According to the structure of the proposed algorithm, it is observed that by
 increasing the number of iterations and approaching the actual size of the image,
 the constant $\alpha$ decreases to increase the effect of the $l_0$ norm.
In the last step, to increase the quality of the restored image, depending on the
type of image,
a non-blind image deblurring algorithm based on the obtained PSF is used.
Also to reduce ringing artifacts,  simple but efficient method based on
Fourier domain restoration filter
and extrapolated image  that is introduced
in \cite{47} is used in the proposed algorithm.
Also in this algorithm the proposed method for threshold of truncating 
gradients in \cite{9} is used
to estimate kernel 
with this difference that fractional gradient values are used instead of the ordinary gradient values.
Numerical results related to the proposed algorithms are studied in the next section.

\noindent\line(1,0){280}\\
\vspace{-0.3cm}
Algorithm 3: Coarse-to-fine framework.\\
\vspace{-0.3cm}
\noindent\line(1,0){280}\\
\begin{algorithmic}
\State \textbf{Input:} Blurred image $y$, the size of PSF, $\lambda$.
\State Obtain $k$ by the coarser level method.
\For{$i=1\rightarrow 5 $}
\State Obtain $x$ by Algorithm \textcolor[rgb]{0.00,0.00,1.00}{1}.
\State Obtain $k$ by Algorithm \textcolor[rgb]{0.00,0.00,1.00}{2}.
\State $\gamma_1\leftarrow \max\{\gamma_1/1.1,10e-4\}$.
\State $\alpha\leftarrow \max\{\alpha/1.1,10e-4\}$.
\EndFor
\State \textbf{Output:} $k$ and intermediate image $x$.
\end{algorithmic}
\line(1,0){280}\\

\section{Experiment results}\label{sec4}
In order to evaluate the performance of the proposed algorithm, different
 types of images as text, face, natural and low-light images are studied in this section.
Also Windows 10-64bit, Intel(R) Core(TM) i3-5005U CPU @2.00GHz, by matlab 2014b
have been used for the calculations.
The results of the algorithm are evaluated using different tests such as
Information content Weighted Structural Similarity Measure
(IW-SSIM)\cite{49}, Multi-scale Structural Similarity (M-SSIM) \cite{50}, Feature Structural Similarity(F-SSIM) \cite{51}
and Peak Signal-to-Noise Ratio (PSNR).\\

\noindent\textbf{Levin dataset:} As a first example, we use Levin dataset. The results are given in Table \ref{Tab1}.
In this table the first number in parentheses
stands for number of image and the second number denotes the number of kernel
in dataset.  
The results in this table are compared with the results in \cite{11,48}, 
and although in some
cases the results of the proposed method have lower values compared to 
these methods, but in general the average of the proposed method 
has a better output compared to these methods. 

	\begin{center}
		\begin{table}[H]
				\begin{threeparttable}
					\scriptsize
					\caption{{\footnotesize Results for Levin dataset.}}
					\label{Tab1}
					\centering
					\begin{tabular}{|@{\hskip2pt}c@{\hskip2pt}|c@{\hskip2pt}c@{\hskip2pt}c@{\hskip2pt}c
							|c@{\hskip2pt}c@{\hskip2pt}c@{\hskip2pt}c|c@{\hskip2pt}c@{\hskip2pt}c@{\hskip2pt}c|}
						\hline
						&\multicolumn{4}{|c|}{method in \cite{11}}   &\multicolumn{4}{c}{method in \cite{48}  }   &\multicolumn{4}{|c|}{proposed method}\\
						(imgae,kernel) &PSNR      &IW-SSIM   &M-SSIM  &F-SSIM      &PSNR     &IW-SSIM &M-SSIM &F-SSIM   &PSNR &IW-SSIM &M-SSIM  &F-SSIM\\
						\hline
						(1,1)         &33.9744    &0.9440    &0.9649  &0.8857    &33.3213  &0.9213  &0.9509 &0.8584&34.0847&0.9464&0.9672&0.8839\\
						(1,2)         &31.8732    &0.6811    &0.8168  &0.7585    &31.8640  &0.7404  &0.8463 &0.7733&31.9585&0.7452&0.8502&0.7699\\
						(1,3)         &32.8360    &0.8487    &0.9114  &0.8291    &32.4375  &0.8054  &0.8861 &0.8122&33.1591&0.8830&0.9307&0.8464\\
						(1,4)         &31.1792    &0.8943    &0.9170  &0.8265    &31.5524  &0.9119  &0.9281 &0.8460&30.8961&0.8587&0.8988&0.7998\\
						(1,5)         &34.2503    &0.9451    &0.9657  &0.8893    &36.4330  &0.9820  &0.9883 &0.9431&34.4087&0.9423&0.9633&0.8773\\
						(1,6)         &31.8843    &0.7944    &0.8760  &0.8031    &32.3136  &0.8351  &0.8998 &0.8198&32.3843&0.8733&0.9202&0.8370\\
						(1,7)         &34.2585    &0.9600    &0.9699  &0.9039    &35.3590  &0.9764  &0.9806 &0.9296&34.6986&0.9706&0.9756&0.9171\\
						(1,8)         &31.0571    &0.8902    &0.9228  &0.8379    &32.0271  &0.9335  &0.9527 &0.8755&31.2542&0.9268&0.9422&0.8708\\
						mean          &32.6641    &0.8697    &0.9181  &0.8417    &33.1635  &0.8882  &0.9291 &0.8573& 32.8555&
						0.8933&    0.9310&    0.8503\\
						\hline
						(2,1)         &31.2466    &0.7454    &0.8206  &0.7363    &31.6891  &0.8299  &0.8750 &0.7803  &32.5778&0.9146&0.9356&0.8461\\
						(2,2)         &30.3489    &0.4617    &0.6317  &0.6892    &30.2430  &0.4208  &0.6092 &0.6903&30.4562&0.4872&0.6551&0.7005\\
						(2,3)         &30.3637    &0.4632    &0.6443  &0.6777    &30.3370  &0.4444  &0.6414 &0.6800&30.4052&0.4767&0.6577&0.6815\\
						(2,4)         &30.7992    &0.8816    &0.8988  &0.8118    &30.8806  &0.8492  &0.8802 &0.7938&30.6572&0.8694&0.8887&0.8021\\
						(2,5)         &32.9145    &0.9093    &0.9328  &0.8485    &32.2432  &0.8434  &0.8850 &0.8124&32.9566&0.9140&0.9357&0.8538\\
						(2,6)         &31.2432    &0.7884    &0.8388  &0.7592    &32.3850  &0.9035  &0.9225 &0.8306&31.6021&0.8193&0.8757&0.7596\\
						(2,7)         &31.3156    &0.8057    &0.8603  &0.7645    &31.1319  &0.7828  &0.8440 &0.7683&31.8457&0.8620&0.8996&0.8009\\
						(2,8)         &31.3195    &0.9265    &0.9392  &0.8596    &31.4929  &0.8911  &0.9173 &0.8279&31.4341&0.9289&0.9415&0.8595\\
						mean          &31.1939    &0.7477    &0.8322  &0.7683    &31.3003  &0.7456  &0.8218 &0.7730&31.4919&0.7840    &0.8487&0.7880\\
						\hline
						(3,1)         &32.8699    &0.9063    &0.9376  &0.8265    &33.6145  &0.9438  &0.9619 &0.8638&33.8982&0.9544&0.9690&0.8841\\
						(3,2)         &31.3196    &0.6660    &0.7868  &0.7026    &31.4307  &0.6967  &0.8030 &0.7102&31.7872&0.7918&0.8621&0.7515\\
						(3,3)         &31.5949    &0.7848    &0.8587  &0.7376    &31.6511  &0.7705  &0.8505 &0.7297&31.8953&0.8285&0.8881&0.7787\\
						(3,4)         &31.3626    &0.8627    &0.9000  &0.7941    &31.9126  &0.8965  &0.9235 &0.8216&31.5777&0.8844&0.9156&0.8117\\
						(3,5)         &34.2074    &0.9522    &0.9680  &0.8779    &35.6246  &0.9801  &0.9864 &0.9250&35.5800&0.9782&0.9852&0.9235\\
						(3,6)         &31.7835    &0.7928    &0.8621  &0.7455    &31.7473  &0.7544  &0.8400 &0.7259&31.6417&0.7841&0.8619&0.7573\\
						(3,7)         &33.7404    &0.9534    &0.9665  &0.8812    &36.3052  &0.9863  &0.9888 &0.9440&34.6544&0.9745&0.9803&0.9197\\
						(3,8)         &31.5923    &0.9412    &0.9553  &0.8683    &31.2153  &0.8531  &0.8973 &0.7831&31.4176&0.9235&0.9437&0.8461\\
						mean          &32.3088    &0.8574    &0.9082  &0.8042    &32.9377  &0.8602  &0.9064 &0.8129& 32.8065&0.8899    &0.9257&0.8341\\
						\hline
						(4,1)         &34.7037    &0.9565    &0.9735  &0.9142    &35.8420  &0.9675  &0.9809 &0.9269&34.6905&0.9546&0.9728&0.9157\\
						(4,2)         &30.2999    &0.5933    &0.7463  &0.7456    &30.5067  &0.5926  &0.7552 &0.7516&30.6186&0.6653&0.7894&0.7717\\
						(4,3)         &30.6522    &0.6810    &0.8011  &0.7630    &31.0039  &0.7400  &0.8366 &0.7896&31.4443&0.78910&0.8700&0.8051\\
						(4,4)         &29.0565    &0.6153    &0.6948  &0.7039    &30.9436  &0.8079  &0.8675 &0.8029&31.3079&0.8121&0.8773&0.8083\\
						(4,5)         &31.3412    &0.7375    &0.8289  &0.7816    &31.1517  &0.6593  &0.7867 &0.7654&32.5577&0.8529&0.9089&0.8328\\
						(4,6)         &31.8929    &0.8596    &0.9093  &0.8392    &32.6546  &0.8786  &0.9207 &0.8539&33.7876&0.9456&0.9644&0.9029\\
						(4,7)         &31.0731    &0.8548    &0.8979  &0.8193    &32.5106  &0.8684  &0.9120 &0.8425&32.7110&0.9075&0.9374&0.8645\\
						(4,8)         &30.0096    &0.8313    &0.8769  &0.8124    &31.8375  &0.9026  &0.9345 &0.8730&31.5005&0.9229&0.9465&0.8836\\
						mean          &31.1286    &0.7662    &0.8411  &0.7974
						&32.0563  &0.8021  &0.8743 &0.8257&32.3273&0.8563    &0.9083&0.8481\\
						\hline
						mean&31.8239  &  0.8103  &  0.8711&    0.8029 &
						32.3644&    0.8240&    0.8829 &   0.8172&
						32.3703&0.8559&0.9034&0.8301\\
						\hline
					\end{tabular}
			\end{threeparttable}
		\end{table}
	\end{center}

\begin{center}
\begin{figure}[h!]
\centering
\makebox[0pt]{
\includegraphics[width=1.0\textwidth]{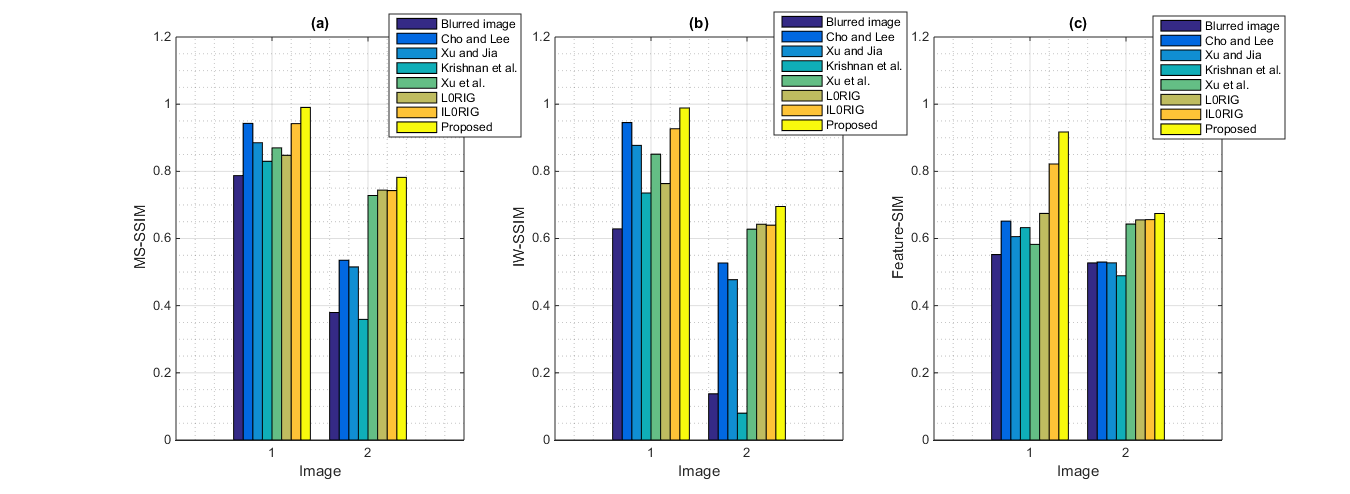}}
\vspace{-0.2cm}
\caption{{\footnotesize
Compare the results for IW-SSIM, M-SSIM and F-SSIM.
} }
\label{fig05}
\end{figure}
\end{center}

\begin{figure}[h!]
  \centering
 \makebox[0pt]{
 	\includegraphics[width=1\textwidth]{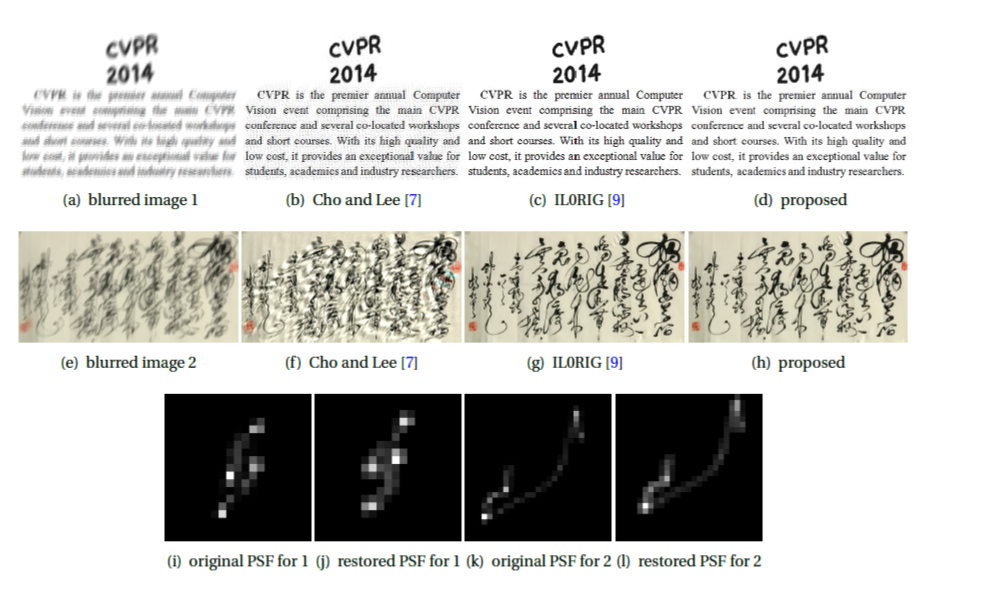}}
  \caption{Visual comparison for text images deblurring.}
  \label{fig06}
\end{figure}

\begin{figure}[h!]
	\centering
\makebox[0pt]{
	\includegraphics[width=1\textwidth]{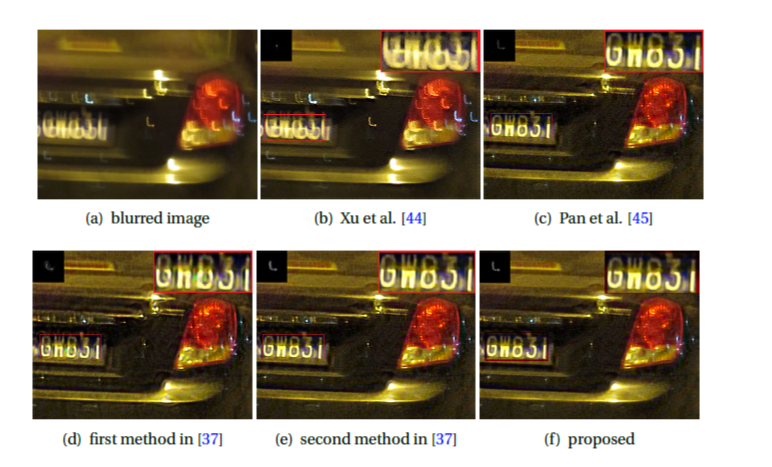}}
	\vspace{-.2cm}
	\caption{Visual comparison for realistic blurred text image.}
	\label{fig07}
\end{figure}

\noindent \textbf{Text image:}
Images containing text usually appear in images obtained by scanning a text.
In recent years, with the expansion of the use of smart phones,
scanning software has expanded \cite{52}. Various factors can affect the quality
of the output image, such as hand shake when using scanning software by smart phone.
The results for two text images such that blurred by kernel 01 and 04 in
Levin dataset are shown in Fig.s \ref{fig05}-\ref{fig07}.
The results of the proposed method are compared with \cite{9,11,48,53,55}.
In Fig. \ref{fig05}, the results for MS-SSIM, IW-SSIM and F-SSIM are given.
Also Fig. \ref{fig06} shows output deblurred images.
The Fig. \ref{fig06} shows the close approximation of the PSF by the proposed algorithm.
Also realistic blurred text image includes car license plate
is restored by the proposed algorithm and compared by methods 
in \cite{48,55,56} in Fig \ref{fig07}.
As can be seen from these figures, the proposed algorithm has a 
better output compared to the other methods.

\begin{figure}[h!]
	\centering
	\makebox[0pt]{
		\includegraphics[width=1\textwidth]{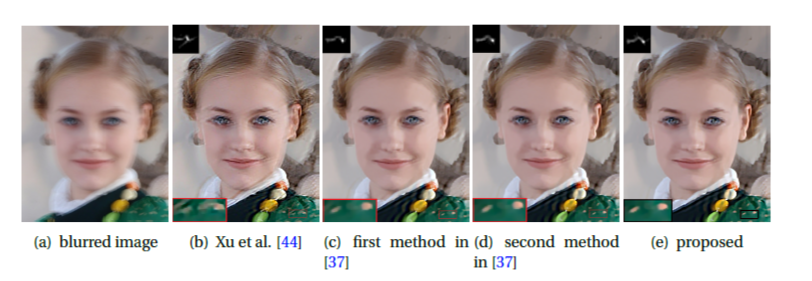}}
	\vspace{-0.2cm}
	\caption{Visual comparison for realistic blurred face image.}
	\label{fig08}
\end{figure}
\newpage

\begin{figure}[H]
\centering
\makebox[0pt]{
\includegraphics[width=1.0\textwidth]{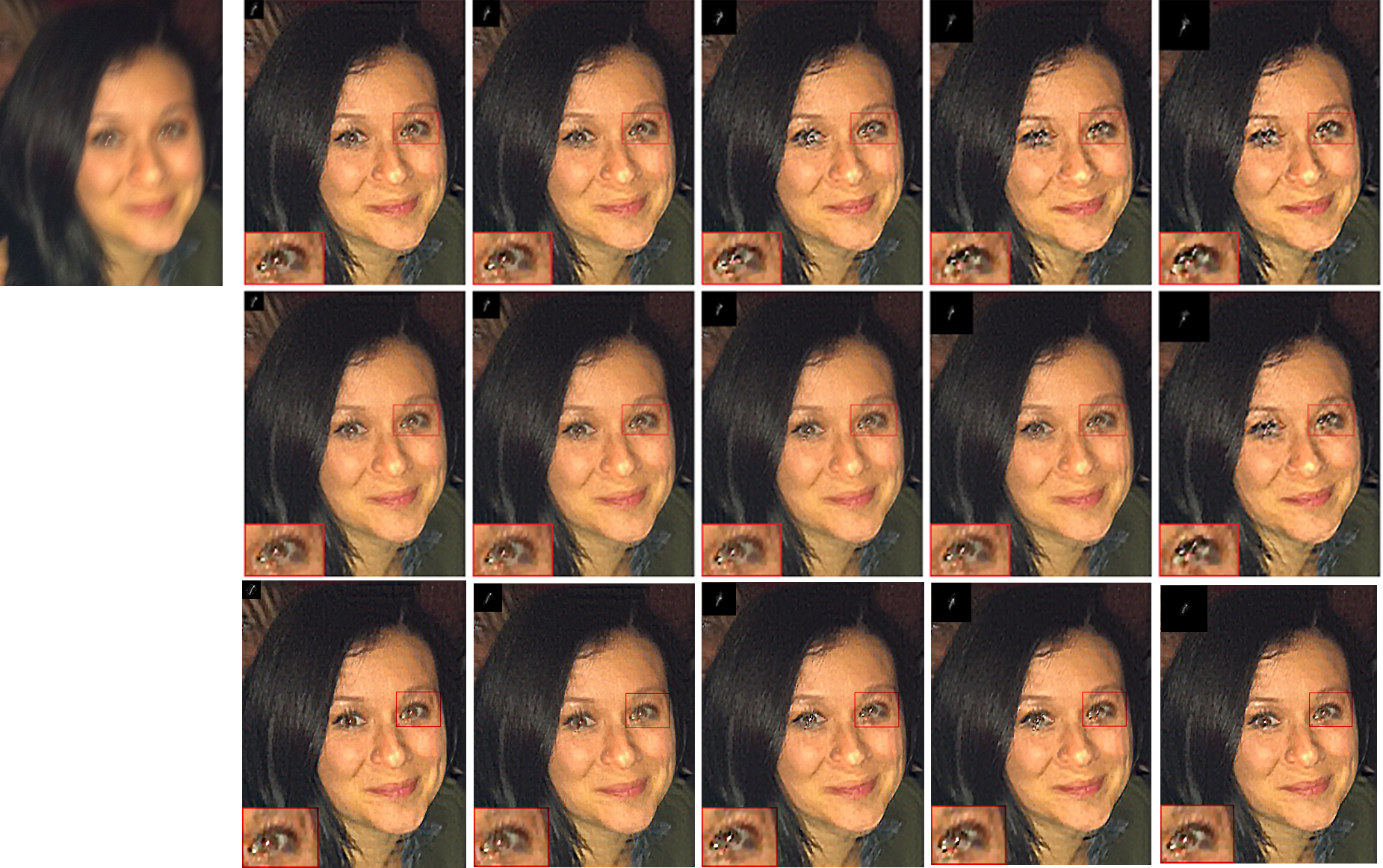}}
\vspace{-.2cm}
\caption{{\footnotesize
Restored face image with the used kernel sizes $\{25, 35, 45, 55, 65\}$, from left to right, respectively.
First row, method in \cite{56}; second row, method in \cite{48}; third row, proposed method.
} }
\label{fig09}
\end{figure}

\noindent \textbf{Face image:}
Another image that is considered in computer science is the face image. This type of image is used in face recognition in topics related to artificial intelligence, so increasing the image quality of the face before use can be considered.
Numerical results related to the proposed method are given in Fig.s \ref{fig08}-\ref{fig09} and are compared with some methods. The results show the efficiency of the proposed algorithm for face images.
The results show that proposed algorithm compares favorably or
even better against compared methods.
It is also seen in Fig. \ref{fig09} that increasing the kernel size in the algorithm, unlike the compared methods, has little effect on the quality of the restored image.

\begin{figure}[h!]
	\centering
\makebox[0pt]{
	\includegraphics[width=1\textwidth]{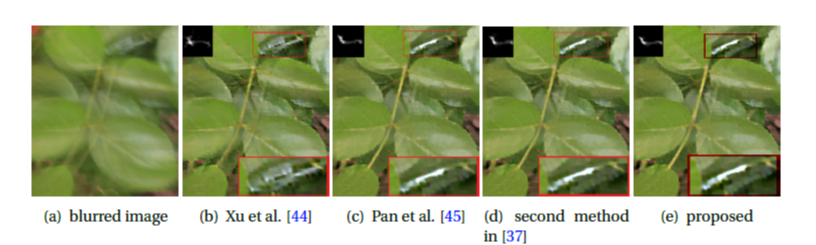}}
	\vspace{-0.2cm}
	\caption{Visual comparison for realistic blurred nature image.}
	\label{fig10}
\end{figure}

\begin{figure}[h!]
	\centering
\makebox[0pt]{
	\includegraphics[width=1\textwidth]{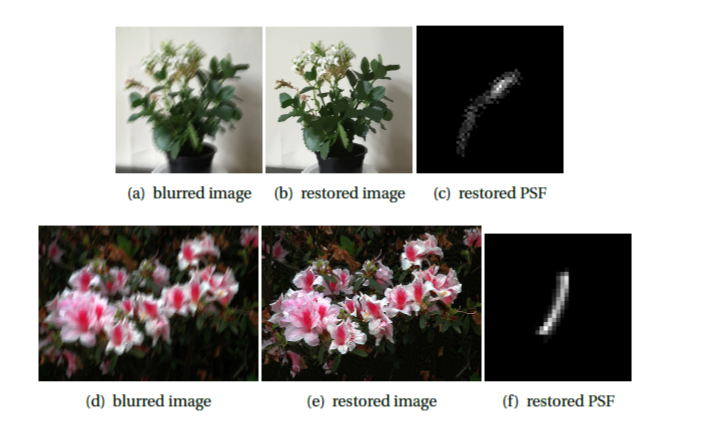}}
	\vspace{-0.2cm}
	\caption{Obtained results by the proposed method for two nature image.}
	\label{fig11}
\end{figure}

\noindent \textbf{Natural image:} Three real natural images are 
studied in this example.
The results are given and compared with method in \cite{48,55,56} 
in Fig.s \ref{fig10} and \ref{fig11}.
The results for this type of images show the efficiency of 
the proposed algorithm in restoring blurred nature images.\\

\begin{figure}[h!]
	\centering
\makebox[0pt]{
	\includegraphics[width=1\textwidth]{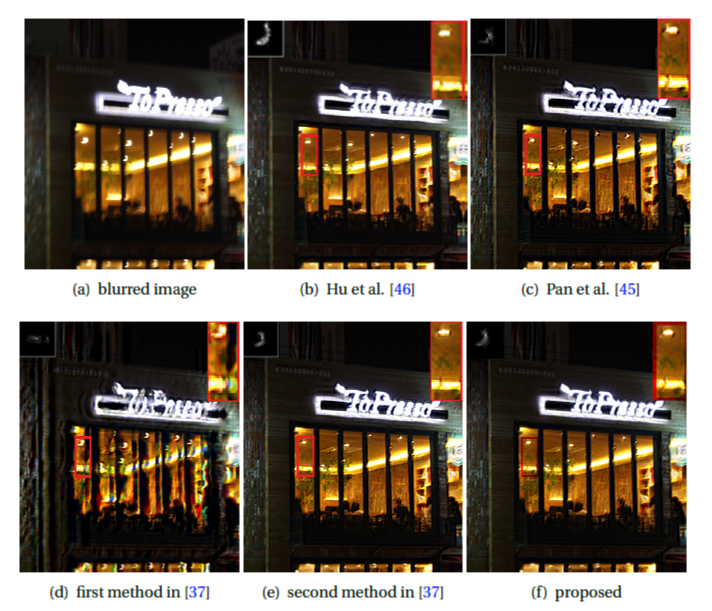}}
\vspace{-.2cm}
	\caption{Visual comparison for realistic blurred low-light image.}
	\label{fig12}
\end{figure}

\noindent \textbf{Low-light image:}
Restoring the low-light image is difficult due to their structure and requires the design of a special algorithm for this type of image \cite{59,60}. However, according to the results of the proposed method in Fig. \ref{fig12}, 
the results show that the proposed algorithm is efficient in recycling this type of images.
In this comparison, we note that the algorithm in \cite{59} is designed specifically for this type of problems.\\

\begin{center}
	\begin{figure}[h!]
		\centering
		\makebox[0pt]{
			\includegraphics[width=1.0\textwidth]{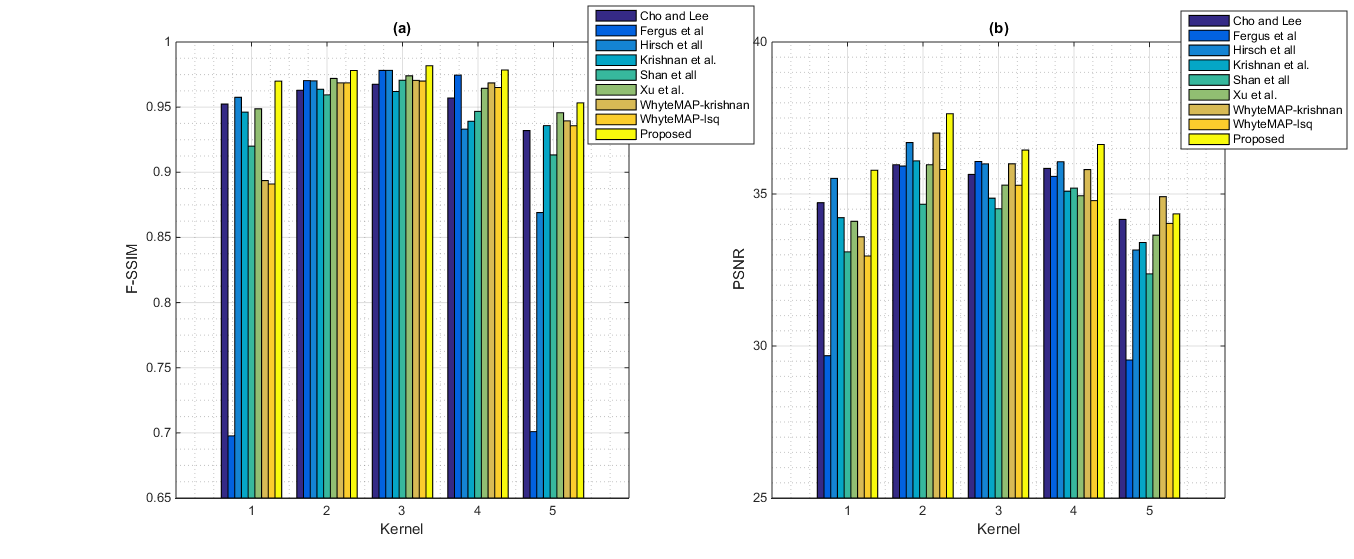}}
		\vspace{-0.2cm}
		\caption{{\footnotesize
				Compare the results for F-SSIM and PSNR with the kernels 1,2,3 and 4.
		} }
		\label{fig14}
	\end{figure}
\end{center}

\begin{figure}[h!]
	\centering
\makebox[0pt]{
	\includegraphics[width=1\textwidth]{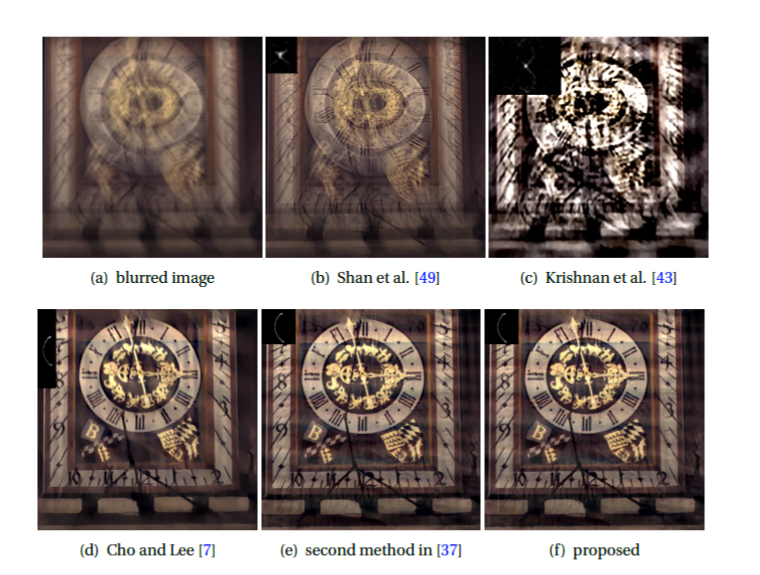}}
	\vspace{-.2cm}
	\caption{Deblurring results of image 1 with kernel 8 in dataset \cite{n59}.}
	\label{fig15}
\end{figure}

\noindent \textbf{Camera motion:} As a final example in this section we review a 
few examples of  real camera motion dataset in \cite{n59} \footnote{See:
\url{https://webdav.tuebingen.mpg.de/pixel/benchmark4camerashake/}}.
The results for this dataset are given in Fig.s \ref{fig14} and \ref{fig15}.
The results of F-SSIM and PSNR values for the church image in this dataset are given in 
Fig. \ref{fig14} and compared with methods in \cite{9,10,54,55,62,63,64}. In this diagram, the restored images of proposed and other methods 
are compared with around 200 ground truth images and the best result is reported as 
the main results. Also the restored images of the clock image with kernel 8 
for the proposed and other algorithms are shown in Fig. \ref{fig15}.\\

\section{Conclusion}\label{sec5}
In this paper, the difference between two norms zero and one is used to approximate the 
point spread function in blind image deblurring problems.
Also in the proposed algorithm, two concepts of framelet and fractional calculations have been used.
The proposed algorithm is evaluated on different types of images with different PSF sizes.
 The results are compared with the other methods and the outputs show 
 the efficiency of the proposed algorithm in deblurring performance.

\end{document}